\pdfoutput=1

\documentclass[11pt]{article}

\usepackage[]{naacl2021}

\usepackage{times}
\usepackage{latexsym}
\usepackage{amsthm,amsmath,amssymb}
\usepackage{multirow}
\usepackage{array}
\usepackage{mathrsfs}
\usepackage{graphicx}
\usepackage{floatrow} 
\usepackage{booktabs}
\usepackage{algorithm}
\usepackage{algorithmic}
\usepackage{amsfonts}
\usepackage{url}
\usepackage{tikz}
\usepackage{tikz-dependency}
\usepackage{pgfplots}
\usepackage{subcaption}
\usetikzlibrary{shapes, positioning, decorations.pathreplacing, shapes.multipart}

\usepackage[T1]{fontenc}

\usepackage[utf8]{inputenc}

\usepackage{microtype}

\newcommand*\samethanks[1][\value{footnote}]{\footnotemark[#1]}

\title{Does syntax matter? A strong baseline for\\ Aspect-based Sentiment Analysis  with RoBERTa}

\author{Junqi Dai\textsuperscript{1,}\thanks{\ \  Equal contribution.} , Hang Yan\textsuperscript{1,}\samethanks \ , Tianxiang Sun\textsuperscript{1}, Pengfei Liu\textsuperscript{2}, Xipeng Qiu\textsuperscript{1,}\thanks{\ \  Corresponding author.} \\
  \textsuperscript{1}Shanghai Key Laboratory of Intelligent Information Processing, Fudan University \\
  \textsuperscript{1}School of Computer Science, Fudan University \\
  \textsuperscript{2}Carnegie Mellon University \\

  \texttt{\{jqdai19,hyan19,txsun19,xpqiu\}@fudan.edu.cn}\\
  \texttt{pliu3@cs.cmu.edu}\\}

\begin{document}
\maketitle

\begin{abstract}
  \textbf{A}spect-\textbf{B}ased \textbf{S}entiment \textbf{A}nalysis (\textbf{ABSA}), aiming at predicting the polarities for aspects, is a fine-grained task in the field of sentiment analysis. Previous work showed syntactic information, e.g.  dependency trees, can effectively improve the ABSA performance.
  Recently, pre-trained models (PTMs) also have shown their effectiveness on ABSA. Therefore, the question naturally arises whether PTMs contain sufficient syntactic information for ABSA so that we can obtain a good ABSA model only based on PTMs. In this paper, we firstly compare the induced trees from PTMs and the dependency parsing trees on several popular models for the ABSA task, showing that the induced tree from fine-tuned RoBERTa (FT-RoBERTa) outperforms the parser-provided tree. The further analysis experiments reveal that the FT-RoBERTa Induced Tree is more sentiment-word-oriented and could benefit the ABSA task. The experiments also show that the pure RoBERTa-based model can outperform or approximate to the previous SOTA performances on six datasets across four languages since it implicitly incorporates the task-oriented syntactic information.\footnote{Our code will be released at \url{https://github.com/ROGERDJQ/RoBERTaABSA}.}

\end{abstract}

\section{Introduction}
\label{sec:intro}
Aspect-based sentiment analysis (ABSA) aims to do the fine-grained sentiment analysis towards aspects~\citep{DBLP:conf/semeval/PontikiGPPAM14,DBLP:conf/semeval/PontikiGPAMAAZQ16}. Specifically, for one or more aspects in a sentence, the  task calls for detecting the sentiment polarities for all aspects. Take the sentence ``great \underline{food} but the \underline{service} was dreadful'' for example, the task is to predict the sentiments towards the underlined aspects, which expects to get polarity \emph{positive} for aspect \emph{food} and polarity \emph{negative} for aspect \emph{service}. Generally, ABSA contains aspect extraction (AE) and aspect-level sentiment classification (ALSC). We only focus on the ALSC task.

Early works of ALSC mainly rely on  manually designed syntactic features, which is labor-intensive yet insufficient. In order to avoid designing hand-crafted features ~\citep{DBLP:conf/acl/JiangYZLZ11,DBLP:conf/semeval/KiritchenkoZCM14}, various neural network models have been proposed in ALSC ~\citep{DBLP:conf/acl/DongWTTZX14,DBLP:conf/ijcai/VoZ15,DBLP:conf/emnlp/WangHZZ16,DBLP:conf/emnlp/ChenSBY17,DBLP:conf/coling/HeLND18,DBLP:conf/sigir/ZhangL019,DBLP:conf/acl/WangSYQW20}.
Since the dependency tree can help the aspects find their contextual words, most of the recently proposed State-of-the-art (SOTA) ALSC models utilize the dependency tree to assist in modeling connections between aspects and their opinion words ~\citep{DBLP:conf/acl/WangSYQW20,DBLP:conf/emnlp/SunZMML19,DBLP:conf/sigir/ZhangL019}. Generally, these dependency tree based ALSC models are implemented in three methods. The first one is to use the topological structure of the dependency tree ~\citep{DBLP:conf/acl/DongWTTZX14,DBLP:conf/emnlp/ZhangLS19,DBLP:conf/emnlp/HuangC19a,DBLP:conf/emnlp/SunZMML19,DBLP:conf/aaai/ZhengZMM20,DBLP:conf/acl/TangJLZ20}; The second one is to use the tree-based distance, which counts the number of edges in a shortest path  between two tokens in the dependency tree ~\citep{DBLP:conf/coling/HeLND18,DBLP:conf/sigir/ZhangL019,DBLP:conf/acl/PhanO20}; The third one is to simultaneously use both the topological structure and the tree-based distance.

Except for the dependency tree, pre-trained models (PTMs)~\citep{qiu2020:scts-ptms}, such as BERT ~\citep{DBLP:conf/naacl/DevlinCLT19}, have also been used to enhance the performance of the ALSC task~\citep{sun2019utilizing, DBLP:conf/acl/TangJLZ20,DBLP:conf/acl/PhanO20,DBLP:conf/acl/WangSYQW20}.
From the view of  interpretability of PTMs, \citet{DBLP:conf/conll/ChenTCCSZ19,DBLP:conf/naacl/HewittM19,DBLP:conf/acl/WuCKL20} try to use probing methods to detect syntactic information in  PTMs. Empirical results reveal that PTMs capture some kind of dependency tree structures  implicitly.

Therefore, two following questions arise naturally.

\textbf{Q1: Will the tree  induced from PTMs achieve better performance than the tree given by a dependency parser when combined with different tree-based ALSC models?}
To answer this question, we choose one model from each of the three typical dependency tree based methods in ALSC, and compare their performance when combined with the parser-provided dependency tree and the  off-the-shelf PTMs induced trees.

\textbf{Q2: Will PTMs adapt the implicitly entailed tree structure to the ALSC task during the fine-tuning?}
Therefore, in this paper, we not only use the trees  induced from the off-the-shelf PTMs to enhance ALSC models, but also use the trees induced from the fine-tuned PTMs~(In short FT-PTMs) which are fine-tuned on the ALSC datasets. Experiments show that trees induced from FT-PTMs can help tree-based ALSC models  achieve better performance than their counterparts before fine-tuning. Besides, models with trees induced from the ALSC fine-tuned RoBERTa can even outperform trees from the dependency parser.

Last but not least, we find that the base RoBERTa with an MLP layer is enough to achieve State-of-the-art (SOTA) or near SOTA performance on all six ALSC datasets across four languages, while incorporating tree structures into RoBERTa-based ALSC models does not achieve concrete improvement.

Therefore, our contributions can be summarized as:

(1) We extensively study the induced trees from  PTMs and FT-PTMs. Experiments show that models using induced trees from FT-PTMs achieve better performance. Moreover, models using induced trees from fine-tuned RoBERTa outperform  other trees.

(2) The analysis of the induced tree  from FT-PTMs shows that it tends to be more sentiment-word-oriented, making the aspect term directly connect to its sentiment adjectives.

(3) We achieve SOTA or near SOTA performances on six ALSC datasets across four languages based on RoBERTa. We find that the RoBERTa could better adapt to ALSC and help the aspects to find the sentiment words.

\section{Related Work}
\textbf{ALSC without Dependencies} \citet{DBLP:conf/ijcai/VoZ15} propose the early neural network model which does not rely on the dependency tree. Along this line, diverse neural network models have been proposed. \citet{DBLP:conf/coling/TangQFL16} use the long short term memory (LSTM) network to enhance the interactions between aspects and context words. In order to model relations of aspects and their contextual words, \citet{DBLP:conf/emnlp/WangHZZ16,DBLP:conf/eacl/ZhangL17,DBLP:conf/ijcai/MaLZW17,DBLP:conf/aaai/TayTH18a} incorporate the attention mechanism into the LSTM-based neural network models. Other model structures such as convolutional neural network (CNN) ~\citep{DBLP:conf/acl/LamLSB18,DBLP:conf/acl/LiX18}, gated neural network ~\citep{DBLP:conf/aaai/ZhangZV16,DBLP:conf/acl/LiX18}, memory neural network ~\citep{DBLP:conf/emnlp/TangQL16,DBLP:conf/emnlp/ChenSBY17, wang-etal-2018-target}, attention neural network ~\citep{tang2019progressive} have also been applied in ALSC.\newline
\textbf{ALSC with Dependencies}  
Early works of ALSC mainly employ traditional text classification methods focusing on machine learning algorithms and manually designed features, which took syntactic structures into consideration from the very beginning.  \citet{DBLP:conf/semeval/KiritchenkoZCM14} combine a set of features including sentiment lexicons and parsing dependencies, from which experiments show the effectiveness of context parsing features.

A myriad of works attempt to fuse dependency tree into neural network models in ALSC.  \citet{DBLP:conf/acl/DongWTTZX14} propose to convert the dependency tree into a binary tree first, then apply the adaptive recursive neural network to propagate information from the context words to aspects. Despite the improvement of aspect-oriented feature modeling, converting the dependency tree into a binary tree might cause syntax related words separated away from each other. In general, owing to the syntax parsing errors, early dependency tree based ALSC models  do not show clear preponderance over models without the dependency tree.

However, the introduction of the neural network into the dependency parsing task enhances the parsing quality substantially ~\citep{chen2014fast,DBLP:conf/iclr/DozatM17}. Recent advances,  leveraging graph neural network (GNN) to model the dependency tree~\citep{DBLP:conf/emnlp/ZhangLS19,DBLP:conf/emnlp/HuangC19a,DBLP:conf/emnlp/SunZMML19,DBLP:conf/acl/TangJLZ20,DBLP:conf/acl/WangSYQW20}, have achieved significant performance.  Among them, \citet{DBLP:conf/aaai/ZhengZMM20,DBLP:conf/acl/WangSYQW20} attempt to convert the dependency tree into the aspect-oriented dependency tree. Instead of using the topological structure of dependency tree, \citet{DBLP:conf/coling/HeLND18,DBLP:conf/sigir/ZhangL019,DBLP:conf/acl/PhanO20} exploit the tree-based distance between two tokens in the dependency tree. \newline
\textbf{PTMs-based Dependency Probing}
Over the past few years, the pre-trained  models (PTMs) have dominated across various NLP tasks. Therefore, many researchers are attracted to investigate what linguistic knowledge has been captured by PTMs~\citep{DBLP:journals/corr/abs-1906-04341,DBLP:conf/emnlp/HewittL19,DBLP:conf/naacl/HewittM19,DBLP:conf/acl/WuCKL20}. \citet{DBLP:journals/corr/abs-1906-04341} try to use a single or a combination of head attention maps of BERT to infer the dependencies. Since BERT has many attention heads, this method can hardly fully reveal the dependency between two tokens. \citet{DBLP:conf/naacl/HewittM19} propose a small learnable probing model to probe the syntax dependencies encoded in BERT. Despite very few parameters been added, it may still be very hard to tell if the syntactic information is encoded by BERT itself or by the additional parameters from the probing model. Therefore, the parameter-free dependency probing method proposed in \citet{DBLP:conf/acl/WuCKL20} might be more preferred.

\section{Method}
In this section, we first introduce how to induce trees from PTMs, then we describe three tree-based ALSC models, which are selected from three representative methods of incorporating the dependency tree in ALSC task.
\subsection{Inducing Tree Structure from PTMs}
Perturbed Masking~\citep{DBLP:conf/acl/WuCKL20} can induce  trees from the pre-trained  models without additional parameters. Generally, a broad range of PTMs can be applied  in the Perturbed Masking method. For the sake of being representative and practical, we select BERT and RoBERTa as our base models.

In this subsection, we first briefly introduce the model structure of BERT and RoBERTa, then present the basic idea of the Perturbed Masking method. More details about them can be found in their respective reference papers.

\subsubsection{BERT and RoBERTa}
BERT~\citep{DBLP:conf/naacl/DevlinCLT19} and RoBERTa~\citep{DBLP:journals/corr/abs-1907-11692} both take Transformers~\citep{DBLP:conf/nips/VaswaniSPUJGKP17} as backbone architecture. Generally, they can be formulated as the following equations
\begin{align}
   & \hat{h^l} = \mathrm{LN}(h^{l-1} + \mathrm{MHAtt}(h^{l-1})), \\
   & h^l = \mathrm{LN}(\hat{h^l} + \mathrm{FFN}(\hat{h^l})),
\end{align}
where $h^0$ is the BERT/RoBERTa input representation,  formed by the sum of token embeddings, position embeddings, and segment embeddings; $\mathrm{LN}$ is the layer normalization layer; $\mathrm{MHAtt}$ is the multi-head self-attention; $\mathrm{FFN}$ contains three layers, the first one is a linear projection layer, then an activation layer, then another linear projection layer; $l$ is the depth of Transformer layers. The base and large version of BERT and RoBERTa have 12, 24 Transformer layers, respectively.

BERT is pre-trained on Masked Language Modeling (MLM) and Next Sentence Prediction (NSP) tasks. In the MLM task, 15\% of the tokens in a sentence are manipulated in three ways. Specifically, 10\%, 10\%, 80\% of them are replaced by a random token, itself, or a ``[MASK]'' token, respectively. In the NSP task, two sentences A and B are concatenated before sending to BERT. Given 50\% of the time when B is the next utterance of A, BERT needs to utilize the vector representation of ``[CLS]'' to figure out whether the input is continuous or not. RoBERTa is only pre-trained on the MLM task.

\subsubsection{Perturbed Masking}
Perturbed Masking aims to detect syntactic information from pre-trained  models.
For a sentence $\mathbf{x}=[x_1, \ldots, x_T]$, BERT and RoBERTa will map each $x_i$ into a contextualized representation $H_{\theta}(\mathbf{x})_i$. Perturbed Masking is trying to derive the value $f(x_i, x_j)$ that denotes the impact a token $x_j$ has on another token $x_i$. To derive this value, it first uses the ``[MASK]'' (or ``$<$mask$>$'' in RoBERTa) to replace the token $x_i$, which returns a representation $H_{\theta}(\mathbf{x} \backslash \{x_i\})_i$ for the masked $x_i$; secondly, it further masks the token $x_j$, which returns a representation $H_{\theta}(\mathbf{x} \backslash \{x_i,x_j\})_i$ with both $x_i,x_j$ being masked. The impact value $f(x_i, x_j)$ is calculated by the Euclidean distance as follows,
\begin{align}
  f\!(\!x_i, \!x_j\!) \!=\! ||\! H_{\theta}(\mathbf{x} \backslash \{x_i\})_i \!-\! H_{\theta}(\mathbf{x} \backslash \!\{\!x_i, \!x_j\!\})_i||_2  
\end{align}
By repeating this process between every two tokens in the sentence, we can get an impact matrix $\mathbf{M} \in \mathbb{R}^{T \times T}$ and $\mathbf{M}_{i,j} = f(x_i, x_j)$. The tree decoding algorithm, such as Eisner~\citep{DBLP:conf/coling/Eisner96} and Chu–Liu/Edmonds' algorithm~{~\citep{1965On,edmonds1967optimum}},  is then used to extract the dependency tree from the matrix $\mathbf{M}$.  The  Perturbed Masking  can exert on any layer of BERT or RoBERTa.

\subsection{ALSC Models Based on Trees}
\label{sec:ALSC_models}
In this subsection, we introduce three representative tree-based ALSC models. Each of the model is from the  methods mentioned in the Introduction part (Section~\ref{sec:intro}). For a fair comparison, all the selected models are of the most recently advanced tree-based ALSC models.
We briefly introduce these three models as follows.

\subsubsection{Aspect-specific Graph Convolutional Networks (ASGCN)}
The Aspect-specific Graph Convolutional Networks (ASGCN) is proposed by \citet{DBLP:conf/emnlp/SunZMML19}. They utilize the dependency tree as a graph, where each word is viewed as a node and the dependencies between words are deemed as edges. After converting the dependency tree into the graph, ASGCN uses the Graph Convolutional Network (GCN) to operate on this graph to model dependencies between each word.
\subsubsection{Proximity-Weighted Convolution Network (PWCN)}
The Proximity-Weighted Convolution Network (PWCN) model is proposed by \citet{DBLP:conf/sigir/ZhangL019}. They try to help the aspect to find their contextual words. For an input sentence, the PWCN first gets its dependency  tree, and based on this tree it would assign a proximity value to each word in the sentence. The proximity value for each word is calculated by the shortest path in the dependency tree between this word and the aspects.
\subsection{Relational Graph Attention Network (RGAT)}
The Relational Graph Attention Network (RGAT) is proposed by \citet{DBLP:conf/acl/WangSYQW20}. In the RGAT model, they transform the dependency tree into an aspect-oriented dependency tree. The aspect-oriented dependency tree uses the aspect as the root node, and all other  words  depend on the aspect directly. The relation between the aspect and other words is either based on the syntactic tag or the tree-based distance in the dependency tree. Specifically, the RGAT reserves syntactic tags for words with 1 tree-based distance to aspect, and assigns virtual tags to longer distance words, such as ``2:con'' for ``A 2 tree-based distance connection''. Therefore, the RGAT model not only exploits the topological structure of the dependency tree but also the tree-based distance between two words.

\section{Experimental Setup}
\label{sec:exp}
In this section, we present details about the datasets, the tree structures used in experiments, as well as the experiments implementations. We conduct experiments  on all six datasets across four languages. But due to the limited space, we present our experiments on the non-English datasets in the Appendix.

\subsection{Datasets}
\label{sec:data}
We run experiments on six benchmark datasets. Three of them, namely, Rest14, Laptop14, and Twitter, are English datasets. Rest14 and Laptop14 are from SemEval 2014 task 4~\citep{DBLP:conf/semeval/PontikiGPPAM14}, containing sentiment reviews from restaurant and laptop domains. Twitter is from \citet{DBLP:conf/acl/DongWTTZX14}, which is processed from tweets.  The statistics of these datasets are presented in Table~\ref{tb:data}. Details of the other three non-English datasets can be found in the Appendix. Following previous works, we remove samples with conflicting polarities or with ``NULL'' aspects in all datasets.

\begin{table}[]
  \centering\small
  \setlength\tabcolsep{1pt}
  \begin{tabular}{m{2.1cm}m{1.1cm}<{\centering}m{1.4cm}<{\centering}m{1.5cm}<{\centering}m{1.4cm}<{\centering}}
    \toprule
    \textbf{Dataset}          & \textbf{Split} & \textbf{Positive} & \textbf{Negative} & \textbf{Neutral} \\\midrule
    \multirow{2}{*}{Rest14}   & Train          & 2164              & 807               & 637              \\
                              & Test           & 728               & 196               & 196              \\ \midrule
    \multirow{2}{*}{Laptop14} & Train          & 994               & 870               & 464              \\
                              & Test           & 341               & 128               & 169              \\\midrule
    \multirow{2}{*}{Twitter}  & Train          & 1561              & 1560              & 3127             \\
                              & Test           & 173               & 173               & 346              \\
    \bottomrule
  \end{tabular}%
  \caption{Data statistics.}
  \label{tb:data}
\end{table}

\subsection{Tree Structures} \label{sec:tree_structure}
For each dataset, we obtain five kinds of trees from three sources. \textbf{(1)} The first one is derived from the off-the-shelf dependency tree parser, such as spaCy\footnote{http://spacy.io/} and allenNLP\footnote{http://www.allennlp.org/}, written as ``Dep.''. For the three English datasets, we use the biaffine parser from the allenNLP package to get the dependency tree, which is reported in \citet{DBLP:conf/acl/WangSYQW20}  that the biaffine parser could achieve better performance.
\textbf{(2)} We induce trees from the pre-trained BERT and RoBERTa by the Perturbed Masking method ~\citep{DBLP:conf/acl/WuCKL20}, written them as ``BERT Induced Tree'' and ``RoBERTa Induced Tree'', respectively.
\textbf{(3)}~We use the  Perturbed Masking method to induce trees from the fine-tuned BERT and RoBERTa after fine-tuning in the corresponding datasets.
These two are written as ``FT-BERT Induced Tree'' and ``FT-RoBERTa Induced Tree''.

Besides, we add ``Left-chain'' and ``Right-chain'' in our experiments. ``Left-chain'', ``Right-chain'' mean that every word deems its previous or next word as the dependent child word.

\newcommand{\tabincell}[2]{\begin{tabular}{@{}#1@{}}#2\end{tabular}}
\begin{table*}[t]
  \centering\small
  \setlength{\tabcolsep}{0pt}
  \begin{tabular}{m{2cm}m{2cm}m{4.75cm}m{1.15cm}<{\centering}m{1.15cm}<{\centering}m{1.15cm}<{\centering}m{1.15cm}<{\centering}m{1.15cm}<{\centering}m{1.15cm}<{\centering}}
    \toprule
    \multicolumn{1}{l}{\multirow{2}{*}{Model}} & \multirow{2}{*}{Tree Features}            & \multirow{2}{*}{Tree Structure} & \multicolumn{2}{c}{Rest14} & \multicolumn{2}{c}{Laptop14} & \multicolumn{2}{c}{Twitter}                                                                       \\
    \cmidrule(r){4-5} \cmidrule(r){6-7} \cmidrule(r){8-9}
    \multicolumn{1}{c}{}                       &                                           &                                 & \centering\textit{Acc.}    & \centering\textit{$F_1$}     & \centering\textit{Acc.}     & \centering\textit{$F_1$} & \centering\textit{Acc.} & \textit{$F_1$} \\
    \midrule
    BiLSTM                                     & -                                         & -                               & 77.59                      & 67.05                        & 70.06                       & 64.46                    & 71.39                   & 69.45          \\
    \midrule
    \multirow{8}{*}{ASGCN}                     & \multirow{8}{*}{\tabincell{l}{Topological                                                                                                                                                                                                   \\  Structure}} & \citet{DBLP:conf/emnlp/ZhangLS19} & 80.86            & 72.19         & 75.55          & 71.05       & 72.15         & 70.40        \\
                                               &                                           & Dep.                            & 81.42                      & 72.87                        & 75.54                       & 71.66                    & 72.36                   & 70.32          \\
    \cmidrule(r){3-3}
                                               &                                           & Left-chain                      & 80.89                      & 71.92                        & 73.98                       & 69.81                    & 71.96                   & 70.47          \\
                                               &                                           & Right-chain\footnotemark[4]     & 80.89                      & 71.92                        & 73.98                       & 69.81                    & 71.96                   & 70.47          \\
    \cmidrule(r){3-3}
                                               &                                           & BERT Induced Tree               & 81.07                      & 72.87                        & 74.29                       & 70.42                    & 72.39                   & 70.25          \\
                                               &                                           & RoBERTa Induced Tree            & 81.16                      & 72.33                        & 74.76                       & 70.0                     & 72.76                   & 71.17          \\
    \cmidrule(r){3-3}
                                               &                                           & FT-BERT Induced Tree            & 81.87                      & 72.89                        & 74.85                       & 70.71                    & 73.36                   & 71.61          \\
                                               &                                           & FT-RoBERTa Induced Tree         & \textbf{82.31}             & \textbf{73.53}               & \textbf{76.33}              & \textbf{72.76}           & \textbf{73.84}          & \textbf{72.66} \\
    \midrule
    \multirow{8}{*}{PWCN}                      & \multirow{8}{*}{\tabincell{l}{Tree-based                                                                                                                                                                                                    \\ Distance}} & \citet{DBLP:conf/sigir/ZhangL019}                             & 80.96            & 72.21         & 76.12          & 72.12       & \multicolumn{1}{c}{-}             & \multicolumn{1}{c}{-}           \\
                                               &                                           & Dep.                            & 80.89                      & 72.16                        & 75.86                       & 71.94                    & 72.10                   & 70.75          \\
    \cmidrule(r){3-3}
                                               &                                           & Left-chain                      & 80.78                      & 72.37                        & 73.35                       & 69.41                    & 71.24                   & 69.42          \\
                                               &                                           & Right-chain\footnotemark[4]     & 80.78                      & 72.37                        & 73.35                       & 69.41                    & 71.24                   & 69.42          \\
    \cmidrule(r){3-3}
                                               &                                           & BERT Induced Tree               & 80.98                      & 72.04                        & 73.82                       & 69.35                    & 72.10                   & 69.90          \\
                                               &                                           & RoBERTa Induced Tree            & 81.16                      & 73.20                        & 73.98                       & 69.94                    & 72.11                   & 70.74          \\
    \cmidrule(r){3-3}
                                               &                                           & FT-BERT Induced Tree            & 81.33                      & 73.57                        & 74.96                       & 70.93                    & 72.54                   & 70.75          \\
                                               &                                           & FT-RoBERTa Induced Tree         & \textbf{82.40}             & \textbf{73.69}               & \textbf{76.95}              & \textbf{73.21}           & \textbf{73.84}          & \textbf{71.43} \\
    \midrule
    \multirow{8}{*}{RGAT}                      & \multirow{8}{*}{\tabincell{l}{Structure                                                                                                                                                                                                     \\ \quad\& \\ Distance}} & \citet{DBLP:conf/acl/WangSYQW20}                             & \textbf{83.30}             & \textbf{76.08}         & 77.42          & 73.76       & \textbf{75.57}         & 73.82       \\
                                               &                                           & Dep.                            & 82.14                      & 74.62                        & 76.49                       & 72.63                    & 74.57                   & 72.57          \\
    \cmidrule(r){3-3}
                                               &                                           & Left-chain                      & 80.53                      & 69.63                        & 74.14                       & 70.04                    & 73.41                   & 71.99          \\
                                               &                                           & Right-chain\footnotemark[4]     & 80.53                      & 69.63                        & 74.14                       & 70.04                    & 73.41                   & 71.99          \\
    \cmidrule(r){3-3}
                                               &                                           & BERT Induced Tree               & 81.27                      & 71.76                        & 75.23                       & 70.47                    & 73.49                   & 72.19          \\
                                               &                                           & RoBERTa Induced Tree            & 81.42                      & 71.79                        & 75.36                       & 71.11                    & 73.78                   & 72.37          \\
    \cmidrule(r){3-3}
                                               &                                           & FT-BERT Induced Tree            & 81.60                      & 72.48                        & 75.96                       & 71.96                    & 74.13                   & 72.47          \\
                                               &                                           & FT-RoBERTa Induced Tree         & 82.76                      & 75.25                        & \textbf{77.43}              & \textbf{74.21}           & 75.43                   & \textbf{74.04} \\
    \bottomrule
  \end{tabular}
  \caption{The  performance(\%)  of tree-based ALSC models incorporating different tree structures on three major English datasets.  Following previous work, Accuracy(\textit{Acc.}) and Marco-$F_1$(\textit{$F_1$}) are used for metric. The reported results are averaged by 3 runs with random initialization. Results named as cited format refer to performance reported in the original paper. Dep. refers to the dependency tree generated from the well-known Biaffine Parser~\citep{DBLP:conf/iclr/DozatM17}. As mentioned in Section~\ref{sec:tree_structure}, BERT Induced Tree, RoBERTa Induced Tree, FT-BERT, and FT-RoBERTa Induced Tree refer to tree structures induced from corresponding PTM. We provide BiLSTM since the other three are different tree-based models over BiLSTM. We highlight the best results of each model in bold.}
  \label{tb:eng}
\end{table*}
\subsection{Implementation Details}
\begin{figure}[]
  \includegraphics[width=1\textwidth]{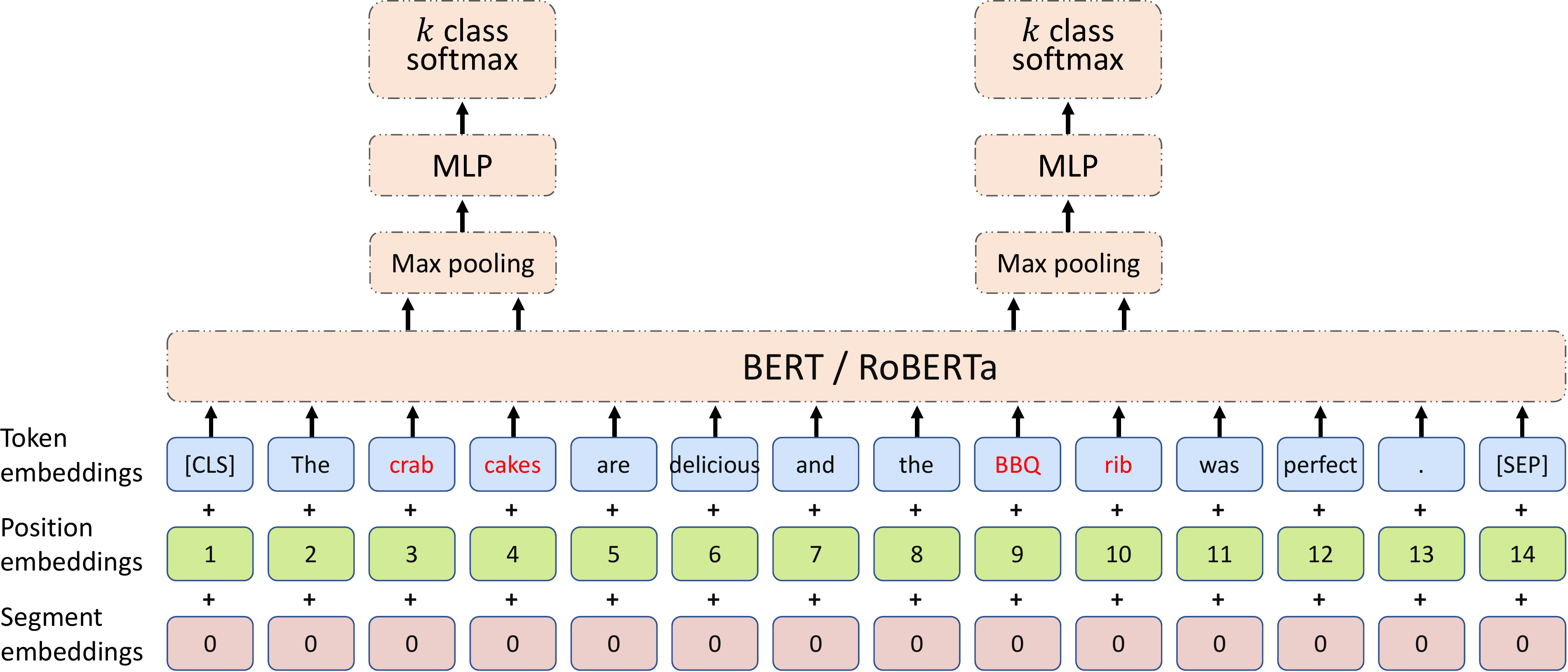}
  \caption{Overall architecture of our fine-tuning model. This structure is enough to achieve SOTA or near SOTA performance in six ALSC datasets  based on RoBERTa.}\label{fig:model_structure}
\end{figure}

In order to derive the FT-PTMs Induced Tree, we  fine-tune BERT and RoBERTa on the ALSC datasets.  To introduce as few parameters as possible, a rather simple MLP is used and the overall structure of our fine-tuning model is presented in Figure~\ref{fig:model_structure}. The fine-tuning experiments are with the batch size $b=32$, dropout rate $d=0.1$, learning rate $\mu=2$e-4 using the AdamW optimizer with the default settings.

As for the Perturbed Masking method, we apply Chu–Liu/Edmonds' algorithm  for the tree decoding. For the induced trees, we first  induce trees from each layer of the PTMs, then test them by the  model in Figure~\ref{fig:model_structure} on dev set which is composed by 20\% of training set. Experiments show that the trees induced from the 11th layer of the PTMs could achieve the best performance among all layers, which is applied for  all  our experiments.

We conduct multiple experiments incorporating different trees (Section~\ref{sec:tree_structure}) into the aforementioned tree-based models (Section~\ref{sec:ALSC_models}). Specifically, we use the 300-dimension Glove~\citep{DBLP:conf/emnlp/PenningtonSM14} embeddings for English datasets.
We keep the word embeddings fixed to avoid overfitting. It is worth noting that in experiments with the RGAT model, since the induced tree does not provide syntactic tags, we assign virtual tags for every dependency in a uniform way, which slightly damage the performance of model.


\section{Experimental Results}
\subsection{ALSC Performance with Different Trees}
\label{sec:pts}
The comparison between models with different trees is presented in Table~\ref{tb:eng}, which comprises experiments results of English datasets. The results of non-English datasets can be found in the Appendix.

We observe that among all the trees, incorporating FT-RoBERTa Induced Tree leads to the best results on all datasets. On average,  models based on the FT-RoBERTa Induced Tree outperform ``Dep.'' by about 1.1\%  in accuracy. This proves the effectiveness and advantage of FT-RoBERTa Induced Tree in this competitive comparison.

\begin{figure*}[bt]
  \ffigbox[\textwidth]
  {
    \begin{subfloatrow}[2]
      \ffigbox[\FBwidth]{
        \includegraphics[width=0.475\textwidth]{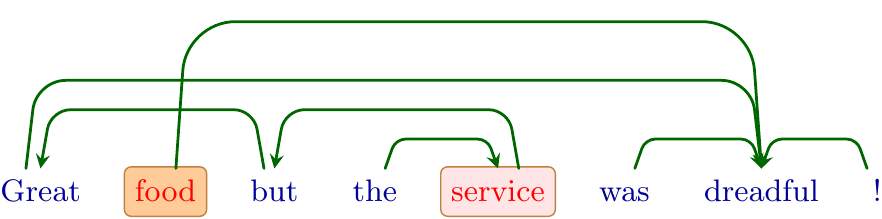}
        \hspace{.15in}
        \includegraphics[width=0.475\textwidth]{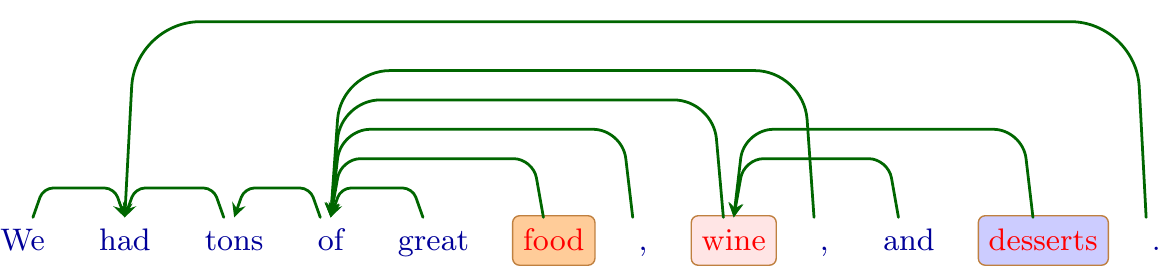}
      }
      {\caption{The parser-provided Tree} \label{case:1}}

    \end{subfloatrow}
    \begin{subfloatrow}[2]
      \ffigbox[\FBwidth]{
        \includegraphics[width=0.475\textwidth]{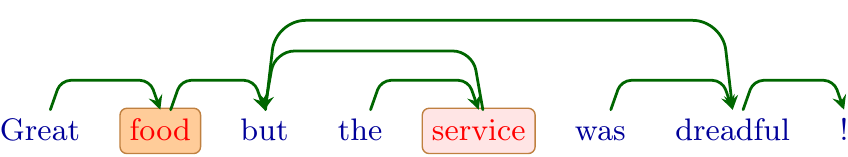}
        \hspace{.15in}
        \includegraphics[width=0.475\textwidth]{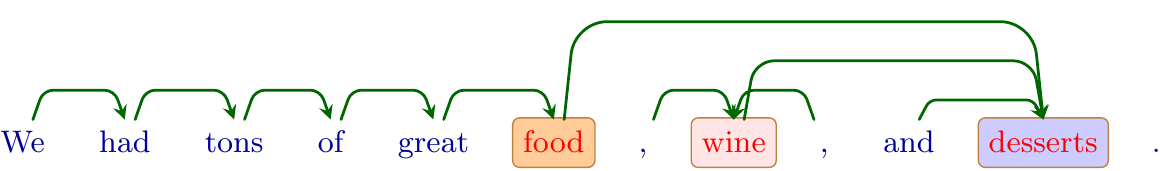}
      }
      {\caption{The RoBERTa Induced Tree} \label{case:2}}

    \end{subfloatrow}
    \begin{subfloatrow}[2]
      \ffigbox[\FBwidth]{
        \includegraphics[width=0.47\textwidth]{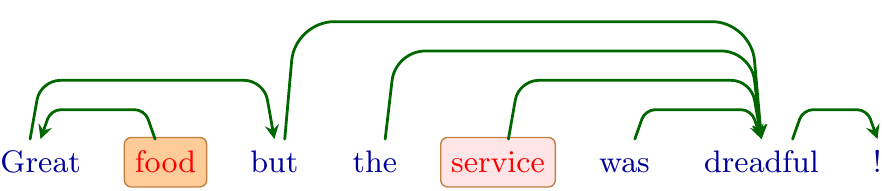}
        \hspace{.15in}
        \includegraphics[width=0.47\textwidth]{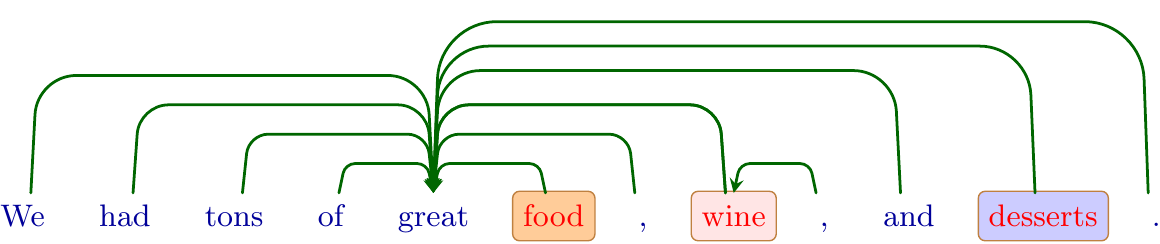}
      }
      {\caption{ The FT-RoBERTa Induced Tree} \label{case:3}}
    \end{subfloatrow}
  }
  {\caption{Visualization of different trees. The colored box refers to the aspect terms. Since ROOT has no directional relation arcs, we omit the  ROOT notation here. For the same two sentences, trees from dependency parser, RoBERTa and fine-tuned RoBERTa are displayed. As Figure~\ref{case:2} shows, trees induced from RoBERTa tend to have more neighboring connections. As the bottom two figures show, trees induced from fine-tuned RoBERTa tend to have connections between sentiment words and others words. } \label{fig:case}}
\end{figure*}

Models using BERT Induced Tree and RoBERTa Induced Tree from Table~\ref{tb:eng} show small performance difference in all but one dataset, and both are close to the ``Left-chain'' and ``Right-chain'' baselines. To have a better sense, we visualize trees induced from RoBERTa in Figure~\ref{case:2}. It shows that RoBERTa Induced Tree has strong neighboring connection dependency pattern. This behavior is expected since the masked language modeling pre-training task will make words favor depending more on its neighboring words. This tendency may be the reason why PTMs induced trees perform similarly to the ``Left-chain'' and ``Right-chain'' baselines.

To answer the question \textbf{Q1} in the Introduction part (Section~\ref{sec:intro}), we need to compare the ``Dep.'', BERT Induced Tree, and RoBERTa Induced Tree  results. The results show that models with dependency trees usually achieve better performance than PTMs induced trees. This is predictable since the word in PTMs induced trees tends to depend on words in their either left or right side as shown in Figure~\ref{fig:case}.
It is worth noting that this observation does not align with the observation in \citet{DBLP:conf/acl/WuCKL20}. The experiments based on PWCN in \citet{DBLP:conf/acl/WuCKL20} show that BERT Induced Tree achieves comparable results with the ``Dep.'', which is consistent with our PWCN results.  However, this observation does not hold when the induced trees are used in a broader range of tree-based ALSC models, especially for the RGAT model in the bottom of Table \ref{tb:eng}. More detailed analysis will be provided in the next section.

Although models with the  PTMs induced trees usually perform worse than those with the dependency parsing trees, models with trees induced from ALSC fine-tuned RoBERTa can surpass both of them. Take RoBERTa Induced Tree and FT-RoBERTa Induced Tree in Table~\ref{tb:eng} for example,  compared with RoBERTa Induced Tree, models incorporating FT-RoBERTa Induced Tree achieves an average accuracy improvement of 1.56\%. This trending is also observed between BERT Induced Tree and FT-BERT Induced Tree.

\floatsetup[subfigure]{heightadjust=object}

\begin{table}[h]
  \centering\small
  \setlength\tabcolsep{1pt}
  \begin{tabular}{m{2.5cm}m{1.5cm}<{\centering}m{2cm}<{\centering}m{1.3cm}<{\centering}}
    \toprule
    Tree Structure & Rest14         & Laptop14       & Twitter        \\
    \midrule
    Dep.           & 0.509          & 0.500          & 0.509          \\
    Left-chain     & 1.000          & 1.000          & 1.000          \\
    Right-chain    & 1.000          & 1.000          & 1.000          \\
    BERT           & 0.710          & 0.690          & 0.741          \\
    RoBERTa        & 0.702          & 0.705          & 0.722          \\
    FT-BERT        & 0.606          & 0.519          & 0.666          \\
    FT-RoBERTa     & \textbf{0.506} & \textbf{0.480} & \textbf{0.485} \\
    \bottomrule
  \end{tabular}
  \caption{Proportion of neighboring connections of different trees in all datasets. We use the short name of induced trees here as well as  Table~\ref{tb:dis} and Table~\ref{tb:baseeng}.}
  \label{tb:1hop}
\end{table}

\subsection{Analysis}
\footnotetext[4]{The Left/Right-chain are exactly the same input files after the data preprocessing in these three models.}
To  further investigate the reasons for the difference between trees, we  propose a set of quantitative metrics, presented in Table~\ref{tb:1hop} and Table~\ref{tb:dis}.

The \textbf{ Proportion of Neighboring Connections} is to calculate the proportion of neighboring connections in the sentence, shown in Table~\ref{tb:1hop}. A neighboring connection links the word to its left/right neighbor word. From Table~\ref{tb:1hop}, we observe that on average over 70\%  relations in BERT/RoBERTa Induced Tree are neighboring connections. This will damage the performance of models using topological structures of trees. Thus, PTMs induced trees usually perform worse than ``Dep.'', with a slight improvement over left/right-chains.


In comparison with RoBERTa Induced Tree, a significant decline of the proportion is shown in FT-RoBERTa Induced Tree  in Table~\ref{tb:1hop}. We see the same tendency in BERT Induced Tree and FT-BERT Induced Tree.  This marks the consistent structure change in the fine-tuning process, indicating the transition to a more diverse structure. As shown in Figure~\ref{case:2}, RoBERTa Induced Tree  has a clear pattern to depend on words in their neighbor side. Yet FT-RoBERTa Induced Tree in Figure~\ref{case:3} shows a more diverse dependency pattern.

\textbf{Aspects-sentiment Distance} is the average distance between aspect and sentiment words. We pre-define a sentiment words set $C$. For a sentence $S_i$ in datasets $S$,  the set of aspects words in $S_i$ is termed as $w$. $S_{i} \cap C $ is the set of sentiment words appearing both in the sentence $S_{i}$ and the sentiment words set $C $. The Aspects-sentiment Distance(AsD) is calculated as follows:
\begin{align}
   & AsD({S_i}) = \frac{\sum\limits^{w_i}_w\sum\limits^{C'_i}_{C'=S_{i} \cap C }dist(C'_i, w_i)}{\left|w\right| | C'|} \\
   & AsD = \frac{\sum\limits^{S_i}_S AsD({S_i})}{|S|}
\end{align}
where $\!|\!\cdot\!|$ is the number of elements in the set and $dist(x_i,x_j)$ represents the relative distance between $x_i$ and $x_j$ in the tree. Specifically,  $C$ contains sentiment words counted on Amazon-2 from \citet{DBLP:conf/acl/TianGXLHWWW20}, which can be found in the Appendix. As for the Rest14 and Laptop14, \citet{dblp:conf/emnlp/xullb20} provides the paired sentiment words with its corresponding aspect. We also calculate the paired Aspects-sentiment Distance(pAsD) on these two datasets, which only counts the distance between aspect and its corresponding sentiment words.
\begin{table}[h]
  \centering\small
  \setlength\tabcolsep{1pt}
  \begin{tabular}{m{2.2cm}m{1.65cm}<{\centering}m{2cm}<{\centering}m{1.3cm}<{\centering}}
    \toprule
    Tree Structure & Rest14                 & Laptop14               & Twitter       \\
    \midrule
    Dep.           & 4.46 / 3.19            & 3.77 / 3.13            & 4.26          \\
    Left-chain     & 7.49 / 6.06            & 6.48 / 5.97            & 7.90          \\
    Right-chain    & 7.49 / 6.06            & 6.48 / 5.97            & 7.90          \\
    BERT           & 5.85 / 4.20            & 5.06 / 4.19            & 5.87          \\
    RoBERTa        & 5.05 / 3.61            & 4.49 / 3.67            & 5.39          \\
    FT-BERT        & 3.85 / 3.58            & 3.65 / 3.22            & 5.06          \\
    FT-RoBERTa     & \textbf{3.56 / 2.92  } & \textbf{3.35 / 2.88  } & \textbf{3.55} \\
    \bottomrule
  \end{tabular}
  \caption{The Aspects-sentiment Distance of different trees in all datasets. The less result indicates shorter distance between aspects and sentiment words. The values of Rest14 and Laptop14 are formed like ``pAsD / AsD''. }
  \label{tb:dis}
\end{table}

\begin{table*}[!ht]
  \setlength{\tabcolsep}{0pt}\centering\small
  \begin{tabular}{m{3cm}m{3.3cm}m{2.5cm}m{1.15cm}<{\centering}m{1.15cm}<{\centering}m{1.15cm}<{\centering}m{1.15cm}<{\centering}m{1.15cm}<{\centering}m{1.15cm}<{\centering}}
    \toprule
    \multirow{2}{*}{Embedding}        & \multirow{2}{*}{Model}                    & \multirow{2}{*}{Tree Structure} & \multicolumn{2}{c}{Rest14}        & \multicolumn{2}{c}{Laptop14}       & \multicolumn{2}{c}{Twitter}                                                                                                                     \\
    \cmidrule(r){4-5} \cmidrule(r){6-7} \cmidrule(r){8-9}
                                      &                                           &                                 & \multicolumn{1}{c}{\textit{Acc.}} & \multicolumn{1}{c}{\textit{$F_1$}} & \multicolumn{1}{c}{\textit{Acc.}} & \multicolumn{1}{c}{\textit{$F_1$}} & \multicolumn{1}{c}{\textit{Acc.}} & \multicolumn{1}{c}{\textit{$F_1$}} \\
    \midrule
    \multirow{4}{*}{Static Embedding} & \tabincell{l}{\textbf{BiLSTM} $^\dagger$} & -                               & 77.59                             & 67.05                              & 70.06                             & 64.46                              & 71.39                             & 69.45                              \\
                                      & \textbf{LSTM+SynATT $^\sharp$      }      & Dep.                            & 80.45                             & 71.26                              & 72.57                             & 69.13                              & \multicolumn{1}{c}{-}             & \multicolumn{1}{c}{-}              \\
                                      & \textbf{AdaRNN $^\sharp$           }      & Dep.                            & \multicolumn{1}{c}{-}             & \multicolumn{1}{c}{-}              & \multicolumn{1}{c}{-}             & \multicolumn{1}{c}{-}              & 66.30                             & 65.90                              \\
                                      & \textbf{TD-GAT $^\sharp$           }      & Dep.                            & 80.35                             & 76.13                              & 74.13                             & 72.01                              & 72.68                             & 71.15                              \\
    \midrule
    \multirow{4}{*}{BERT}             & \textbf{MLP          }                    & -                               & 85.35                             & 78.38                              & 78.36                             & 74.16                              & 75.92                             & 74.41                              \\
                                      & \textbf{DGEDT $^\sharp$         }         & Dep.                            & 86.30                             & 80.0                               & 79.80                             & 75.60                              & \textbf{77.90}                    & 75.40                              \\
                                      & \textbf{RGAT  $^\sharp$       }           & Dep.                            & 86.60                             & 81.35                              & 78.21                             & 74.07                              & 76.15                             & 74.88                              \\
                                      & \textbf{RACL$^\sharp$         }           & -                               & \multicolumn{1}{c}{-}             & \textbf{81.61}                     & \multicolumn{1}{c}{-}             & 73.91                              & \multicolumn{1}{c}{-}             & 81.61                              \\
    \midrule
    \multirow{9}{*}{RoBERTa}          & \textbf{MLP          }                    & -                               & 87.37                             & 80.96                              & 83.78                             & 80.73                              & 77.17                             & \textbf{76.20}                     \\
                                      & \textbf{RoBERTa-ASC $^\sharp$ }           & Dep.                            & 82.82                             & 75.12                              & 74.12                             & 70.52                              & \multicolumn{1}{c}{-}             & \multicolumn{1}{c}{-}              \\
                                      & \textbf{LCFS-ASC-CDW $^\sharp$}           & Dep.                            & 86.71                             & 80.31                              & 80.52                             & 77.13                              & \multicolumn{1}{c}{-}             & \multicolumn{1}{c}{-}              \\
                                      & \textbf{ASGCN        }                    & Dep.                            & 86.90                             & 80.75                              & 81.66                             & 78.31                              & 75.28                             & 74.38                              \\
                                      &                                           & \tabincell{l}{FT-RoBERTa}       & 86.87                             & 80.59                              & 83.33                             & 80.32                              & 76.10                             & 75.07                              \\
                                      & \textbf{PWCN         }                    & Dep.                            & 87.41                             & 81.07                              & \textbf{84.16}                    & \textbf{81.18}                     & 76.63                             & 75.60                              \\
                                      &                                           & \tabincell{l}{FT-RoBERTa}       & 87.35                             & 80.85                              & 84.01                             & 81.08                              & 77.02                             & 75.52                              \\
                                      & \textbf{RGAT         }                    & Dep.                            & 87.43                             & 80.61                              & 83.43                             & 80.28                              & 74.42                             & 72.93                              \\
                                      &                                           & \tabincell{l}{FT-RoBERTa}       & \textbf{87.52}                    & 81.29                              & 83.33                             & 79.95                              & 75.81                             & 74.91                              \\
    \bottomrule
  \end{tabular}
  \caption{The results(\%) of SOTA ALSC models on English datasets. The results with ``$\dagger$'' are retrieved from \citet{DBLP:conf/emnlp/SunZMML19}, and those with ``$\sharp$'' are retrieved from the original papers. Those without additional symbols are on our own. We highlight the best  results on bold. }
  \label{tb:baseeng}
\end{table*}

We present the Aspects-sentiment Distance~(AsD) of different trees in English datasets in Table~\ref{tb:dis}. Results show that FT-RoBERTa has the least AsD value, indicating the shortest aspects-sentiment distance. Compared to PTMs induced trees, the trees from  FT-PTMs have less AsD, indicating shortened aspects-sentiment distance. This shows that the FT-PTMs induced trees are more sentiment-word-oriented, which partially reveals that the fine-tuning in ALSC encourages the aspects to find sentiment words. However, for the ``Dep.'', we notice that some Twitter results in Table~\ref{tb:eng} can not be fully explained  by these two proposed metrics. We conjecture that the grammar casualness  features the Twitter corpus, which makes the parser hard to provide an accurate dependency parsing tree. Still, these two metrics can be suitable for the induced trees.

Taken together, as the conclusion to \textbf{Q2}, these analyses demonstrate that the fine-tuning on ALSC could adapt the induced tree implicitly. On the one hand, less proportion of neighboring connections after fine-tuning indicates the increase of long range connections. On the other hand, less Aspects-sentiment Distance  after fine-tuning illustrates the shorter distance between aspects and sentiment words, which helps to model connections between aspects and sentiment words.
Thus, as shown in Section~\ref{sec:pts}, fine-tuning RoBERTa in ALSC not only makes induced tree better suit the ALSC task but also outperform the dependency tree when combined with different tree-based ALSC models.


\subsection{Comparison between ALSC models}
\label{sec:ana}
Additional, we explore how well the fine-tuned RoBERTa model could achieve in the ALSC task. We select a set of top high-performing models of ALSC as state-of-the-art alternatives. The comparison results are shown in Table~\ref{tb:baseeng}.

Comparing with all these SOTA alternatives, surprisingly, the RoBERTa with an MLP layer achieve SOTA or near SOTA performance. Especially, compared to other datasets, we notice that significant improvement is obtained on the Laptop14 dataset. We assume that the pre-training corpus of RoBERTa may be more friendly to the laptop domain since the RoBERTa-MLP already obtains much better results than the  BERT-MLP on Laptop14.  For these BERT-based models in the second row of Table~\ref{tb:baseeng},  similar experiments using RoBERTa are conducted. However, limited improvements have been made over the RoBERTa-MLP. We expect that induced trees from models specifically pre-trained for ALSC~\citep{DBLP:conf/acl/TianGXLHWWW20} may provide more information, which is left for the future works.


The FT-RoBERTa Induced Tree could be beneficial to Glove based ALSC models. However, incorporating trees over the RoBERTa brings no significant improvement, even the decline can be seen in some cases. This may be caused by failure to reconcile  the implicitly entailed tree with external tree. We argue that incorporating trees over the RoBERTa in currently widely-used tree methods    may be the loss outweighs the gain.
Additionally, in the review of previous ALSC works, we notice that very few works employ the RoBERTa as the base model. We would attribute this to the difficulty of  optimizing the RoBERTa-based ALSC models. As the higher architecture, which is usually randomly initialized, needs a bigger learning rate compared to the RoBERTa. The  inappropriate hyperparameters may be  the cause reason for the lagging performance of previous RoBERTa-based ALSC works~\citep{DBLP:conf/acl/PhanO20}.

\section{Conclusion}
In this paper, we analyze several tree structures for the ALSC task including parser-provided dependency tree and PTMs-induced tree. Specifically, we induce trees  using the Perturbed Masking method from the original PTMs and ALSC fine-tuned PTMs respectively, and then compare the different tree structures on three typical tree-based ALSC models on six datasets across four languages.  Experiments reveal that fine-tuning on ALSC task forces PTMs to implicitly learn more sentiment-word-oriented trees, which can bring  benefits to Glove based ALSC models. Benefited from its better implicit syntactic information, the fine-tuned RoBERTa with an MLP is enough to obtain SOTA or near SOTA results for ALSC task.
Our work can lead to  several promising directions, such as PTMs-suitable tree-based models and better tree-inducing methods from PTMs.

\section*{Acknowledgment}
We would like to thank the anonymous reviewers for their helpful comments. This work was supported by the National Key Research and Development Program of China (No. 2020AAA0106700) and  National Natural Science Foundation of China (No. 62022027).






\bibliography{anthology,naacl2021}
\bibliographystyle{acl_natbib}

\appendix
\section{Experiments on non-English Datasets}
In this section, we provide details about our experiments on non-English datasets.
\subsection{Datasets}

We conduct experiments on  three non-English datasets, which are named Dutch, French, and Spanish, respectively. All of them are restaurant review datasets from SemEval-2016 task 5~\citep{DBLP:conf/semeval/PontikiGPAMAAZQ16}, whose languages are the same as dataset names. Detailed  data statistics can be found in Table~\ref{tb:data}. Following previous works, we remove samples with conflicting polarities or with ``NULL'' aspect terms in all datasets.

\begin{table}[h]
  \centering\small
  \setlength\tabcolsep{1pt}
  \begin{tabular}{m{2.1cm}m{1.1cm}<{\centering}m{1.4cm}<{\centering}m{1.5cm}<{\centering}m{1.4cm}<{\centering}}
    \toprule
    \textbf{Dataset}         & \textbf{Split} & \textbf{Positive} & \textbf{Negative} & \textbf{Neutral} \\\midrule
    \multirow{2}{*}{Dutch}   & Train          & 720               & 386               & 108              \\
                             & Test           & 229               & 120               & 23               \\\midrule
    \multirow{2}{*}{French}  & Train          & 833               & 683               & 98               \\
                             & Test           & 320               & 253               & 54               \\\midrule
    \multirow{2}{*}{Spanish} & Train          & 1308              & 443               & 79               \\
                             & Test           & 505               & 171               & 33               \\
    \bottomrule
  \end{tabular}%
  \caption{Data statistics.}
  \label{tb:data}
\end{table}

\subsection{Tree Structures}
We obtain five kinds of trees for every dataset. The first one is to use the off-the-shelf dependency tree parser to get parser-provided dependency trees, written as “Dep.”. Specifically,  we utilize the spaCy parser for the non-English datasets. The second method is to induce the trees from the pre-trained mBERT and XLM-R~\citep{DBLP:conf/acl/ConneauKGCWGGOZ20} base models by the Perturbed Masking method ~\citep{DBLP:conf/acl/WuCKL20}, written them as ``BERT Induced Tree'' and ``RoBERTa Induced Tree'', respectively. The third method is to use the same method as above to induce trees from the mBERT and XLM-R after fine-tuning in the corresponding datasets with the same model structure as English datasets. These two are written as ``FT-BERT Induced Tree'' and ``FT-RoBERTa Induced Tree'' to have a uniform form as the English datasets.  Similarly,  we add ``Left-chain'' and ``Right-chain'' as baselines. ``Left-chain'', ``Right-chain'' mean that every word deems its previous or next word as the dependent child word.
\subsection{Implementation Details}
Similar to the English datasets, Experiments incorporating tree-based ALSC models with different trees are conducted on non-English datasets, as well as the fine-tuning of PLMs. All experiments are conducted on the  NVIDIA GTX1080Ti.

For experiments with tree-based models, we use the 300-dimension pre-trained embeddings~\citep{DBLP:conf/emnlp/RuderGB16} for non-English datasets. We keep the word embeddings fixed to avoid overfitting. Other parameters are initialized with original models. It is worth noting that in RGAT Model reproduction, since the induced tree does not provide relation labels, we assign virtual relations for every dependency in a uniform way.

We retain the  fine-tuning experiments with batch size $b=32$, dropout rate $d=0.1$, learning rate $\mu=2$e-4 using the AdamW optimizer with the default settings.

As for the induced trees, We choose the trees induced from the 11th layer in  all of our experiments.

\begin{table*}[tp]
  \centering\setlength{\tabcolsep}{0pt}\small
  \begin{tabular}{m{2cm}m{2cm}m{4.75cm}m{1.15cm}<{\centering}m{1.15cm}<{\centering}m{1.15cm}<{\centering}m{1.15cm}<{\centering}m{1.15cm}<{\centering}m{1.15cm}<{\centering}}
    \toprule
    \multirow{2}{*}{Model} & \multirow{2}{*}{Tree Features}            & \multirow{2}{*}{Tree Sturcture} & \multicolumn{2}{c}{Dutch} & \multicolumn{2}{c}{French} & \multicolumn{2}{c}{Spanish}                                                                       \\
    \cmidrule(r){4-5}\cmidrule(r){6-7}\cmidrule(r){8-9}
                           &                                           &                                 & \centering\textit{Acc.}   & \centering\textit{$F_1$}   & \centering\textit{Acc.}     & \centering\textit{$F_1$} & \centering\textit{Acc.} & \textit{$F_1$} \\
    \midrule
    BiLSTM                 & -                                         & -                               & 83.30                     & 62.50                      & 80.0                        & 67.50                    & 85.30                   & 62.10          \\
    \midrule
    \multirow{5}{*}{ASGCN} & \multirow{5}{*}{\tabincell{l}{Topological                                                                                                                                                                                                \\  Structure}}  & Dep.                            & 84.18         & 70.06       & 79.23         & 65.0          & 87.6          & 67.36       \\
                           &                                           & BERT Induced Tree               & 84.45                     & 67.25                      & 79.23                       & 66.31                    & 87.10                   & 67.58          \\
                           &                                           & FT-BERT Induced Tree            & 83.37                     & 68.12                      & 79.38                       & 62.27                    & 86.70                   & 69.07          \\
                           &                                           & RoBERTa Induced Tree            & 84.45                     & \textbf{70.94}             & 79.53                       & 67.20                    & 86.70                   & 68.19          \\
                           &                                           & FT-RoBERTa Induced Tree         & \textbf{84.99}            & 68.26                      & \textbf{80.31}              & \textbf{67.4 }           & \textbf{87.8 }          & \textbf{72.88} \\
    \midrule
    \multirow{5}{*}{PWCN}  & \multirow{5}{*}{\tabincell{l}{Relative                                                                                                                                                                                                   \\ Distance}} & Dep.                            & 83.38         & 67.82       & 79.23         & 66.28       & 86.25         & \textbf{67.95}       \\
                           &                                           & BERT Induced Tree               & 84.18                     & 67.37                      & 78.46                       & 64.6                     & 87.09                   & 66.57          \\
                           &                                           & FT-BERT Induced Tree            & 84.18                     & 68.17                      & 78.62                       & 66.57                    & 86.53                   & 67.87          \\
                           &                                           & RoBERTa Induced Tree            & 84.90                     & 68.30                      & 78.62                       & 63.27                    & 85.97                   & 66.38          \\
                           &                                           & FT-RoBERTa Induced Tree         & \textbf{85.25}            & \textbf{70.21}             & \textbf{80.0 }              & \textbf{67.9 }           & \textbf{87.23}          & 64.93          \\
    \midrule
    \multirow{5}{*}{RAGT}  & \multirow{5}{*}{\tabincell{l}{Structure                                                                                                                                                                                                  \\ \quad \& \\ Distance}} & Dep.                            & 84.45         & 59.85       & 79.53         & 66.16       & 86.14         & 56.44       \\
                           &                                           & BERT Induced Tree               & 84.45                     & 57.36                      & 76.92                       & 58.14                    & 86.53                   & 61.70          \\
                           &                                           & FT-BERT Induced Tree            & 84.18                     & 59.67                      & 78.61                       & 60.79                    & 85.50                   & 62.66          \\
                           &                                           & RoBERTa Induced Tree            & 84.71                     & 67.60                      & 78.15                       & 61.10                    & 86.81                   & 61.88          \\
                           &                                           & FT-RoBERTa Induced Tree         & \textbf{85.25}            & \textbf{69.53}             & \textbf{81.38}              & \textbf{66.97}           & \textbf{87.37}          & \textbf{65.30} \\
    \bottomrule
  \end{tabular}
  \caption{The averaged performance(\%) of tree-based ALSC models incorporating different tree structures on three non-English datasets.   Dep. refers to the dependency tree generated by spaCy. As mentioned in English datasets, BERT Induced Tree, RoBERTa Induced Tree, FT-BERT, and FT-RoBERTa Induced Tree refer to tree structures extracted from corresponding PLMs.}
  \label{tb:non_eng}
\end{table*}

\begin{table*}[tp]
  \setlength{\tabcolsep}{0pt}\small
  \begin{tabular}{m{3cm}m{3.3cm}m{2.5cm}m{1.15cm}<{\centering}m{1.15cm}<{\centering}m{1.15cm}<{\centering}m{1.15cm}<{\centering}m{1.15cm}<{\centering}m{1.15cm}<{\centering}}
    \toprule
    \multirow{2}{*}{Embedding}        & \multirow{2}{*}{Model}        & \multirow{2}{*}{Tree Structure} & \multicolumn{2}{c}{Dutch} & \multicolumn{2}{c}{French} & \multicolumn{2}{c}{Spanish}                                                                              \\
    \cmidrule(r){4-5}\cmidrule(r){6-7}\cmidrule(r){8-9}
                                      &                               &                                 & \centering\textit{Acc.}   & \centering\textit{$F_1$}   & \centering\textit{Acc.}     & \centering\textit{$F_1$} & \centering\textit{Acc.} & \textit{$F_1$}        \\
    \midrule\
    \multirow{5}{*}{Static Embedding} & \textbf{BiLSTM}               & -                               & 83.3                      & 62.5                       & 80.0                        & 67.5                     & 85.3                    & 62.1                  \\
                                      & \textbf{SA-LSTM-P $^\sharp$ } & -                               & 87.3                      & \multicolumn{1}{c}{-}      & 82.4                        & \multicolumn{1}{c}{-}    & 88.0                    & \multicolumn{1}{c}{-} \\
                                      & \textbf{Our ASGCN }           & Dep.                            & 81.6                      & 61.0                       & 75.5                        & 63.0                     & 85.0                    & 59.0                  \\
                                      & \textbf{Our RGAT  }           & Dep.                            & 81.0                      & 62.1                       & 75.1                        & 53.3                     & 84.6                    & 55.2                  \\
                                      & \textbf{Our PWCN  }           & Dep.                            & 84.1                      & 69.2                       & 78.4                        & 66.7                     & 86.9                    & 67.5                  \\
    \midrule
    mBERT                             & \textbf{MLP     }             & -                               & 80.37                     & 63.43                      & 78.06                       & 65.04                    & 88.21                   & 68.03                 \\
    \midrule
    \multirow{7}{*}{XLM-R}            & \textbf{MLP}                  & -                               & \textbf{88.36}            & \textbf{76.29}             & 85.95                       & 74.72                    & 91.48                   & 77.96                 \\
                                      & \textbf{ASGCN     }           & Dep.                            & 87.97                     & 74.38                      & 86.43                       & 77.14                    & 91.91                   & 77.49                 \\
                                      &                               & FT-RoBERTa                      & 88.2                      & 75.23                      & 86.04                       & 76.21                    & \textbf{92.47}          & \textbf{78.74}        \\
                                      & \textbf{PWCN      }           & Dep.                            & 88.36                     & 75.72                      & 86.4                        & 76.8                     & 91.51                   & 77.32                 \\
                                      &                               & FT-RoBERTa                      & 88.1                      & 75.54                      & \textbf{86.69}              & \textbf{77.42}           & 91.44                   & 78.13                 \\
                                      & \textbf{RGAT      }           & Dep.                            & 88.31                     & 70.57                      & 85.92                       & 75.14                    & 91.61                   & 76.41                 \\
                                      &                               & FT-RoBERTa                      & 87.86                     & 70.97                      & 86.41                       & 74.38                    & 92.11                   & 76.62                 \\
    \bottomrule
  \end{tabular}
  \caption{The results(\%) of ALSC models incorporating with different tree structures on non-English datasets.  The definition of tree structures retains the same as the aforementioned.  The results with ``$\sharp$'' are retrieved from the original papers.}
  \label{tb:basenoeng}
\end{table*}

\subsection{Experimental Results}
\subsubsection{ALSC Performance with Different Trees}
The comparison between models with different trees is presented in Table~\ref{tb:non_eng}, which comprises experiments results of non-English datasets.
Experimental results shows that:
(1) Incorporating FT-RoBERTa Induced Tree leads to the best results on all datasets, which proves the effectiveness and advantage of FT-RoBERTa Induced Tree in non-English datasets. Moreover, we find that the results of  the FT-RoBERTa Induced Tree usually have more stable $F_1$ scores.
(2) Subjected to the quality of the parser of non-English languages, models using the PLMs induced trees achieve slightly better performance compared to ``Dep.''. This illustrates that the dependency tree could be very sensitive to parser and quality of  corpus.
(3) Similarly, from ``RoBERTa Induced Tree'' and ``FT-RoBERTa Induced Tree'', we  conclude that fine-tuning can substantially enhance the ALSC performance through trees induced from PLMs.


\subsubsection{Comparison between ALSC models}
Similarly, we compare the performance between the fine-tuned XLM-R and a set of top high-performing models. The results are presented in Table~\ref{tb:basenoeng}. We could see that XLM-R with an MLP is enough to achieve SOTA or near SOTA results in non-English datasets.

\section{Sentiment words set}

\begin{table}[h]
  \centering
  \begin{tabular}{|m{2cm}|m{4.5cm}|}
    \hline
    positive sentiment words & great, good, like, just, will, well, even, love, best, better, back, want, recommend, worth, easy, sound, right, excellent, nice, real, fun, sure, pretty, interesting, stars \\
    \hline
    negative sentiment word  & too, little, bad, game, down, long, hard, waste, disappointed, problem, try, poor, less, boring, worst, trying, wrong, least, although, problems, cheap                       \\
    \hline
  \end{tabular}
  \caption{The sentiment words used in our analysis, derived from \citet{DBLP:conf/acl/TianGXLHWWW20}. }
  \label{tb:senti}
\end{table}

To calculate the Aspects-sentiment Distance of different tree structures on English datasets, we pre-define a set of sentiment words, shown in Table~\ref{tb:senti}. Specifically, we use the sentiment words described in \citet{DBLP:conf/acl/TianGXLHWWW20}, which are the selected 50 most frequent sentiment words counted on Amazon-2.

\end{document}